\documentclass{article}

\usepackage[final, nonatbib]{neurips_2019}

\usepackage[utf8]{inputenc} 
\usepackage[T1]{fontenc}    
\usepackage{hyperref}       
\usepackage{url}            
\usepackage{booktabs}       
\usepackage{amsfonts}       
\usepackage{nicefrac}       
\usepackage{microtype}      
\usepackage{float}
\usepackage{array}
\usepackage{amsmath}
\usepackage{amsfonts}
\usepackage{amssymb}
\usepackage{amsthm}
\usepackage{graphicx}
\usepackage{multirow}
\usepackage{booktabs}
\usepackage{xcolor}
\usepackage{listings}
\usepackage{datetime}
\usepackage{hyperref}
\usepackage{csquotes}

\newcommand{\obs}{Y} 
\newcommand{\ex}[1]{\mathbb{E}\left[ #1 \right]}  
\newcommand{\IR}{\mathbb{R}}  
\newcommand{\eg}{\emph{e.g.}}  
\newcommand{\ie}{\emph{i.e.}}  

\title{Intermittent Demand Forecasting with Deep Renewal Processes}

\author{%
  Ali Caner T\" urkmen \\
  Department of Computer Engineering\\
  Bo\u gazi\c ci University \\
  Istanbul, Turkey\\
  \texttt{caner.turkmen@boun.edu.tr}\\
  \And
  Yuyang Wang \\
  AWS AI Labs \\
  2100 University Ave \\
  East Palo Alto, CA, US\\
  \texttt{yuyawang@amazon.com} \\
  \And
  Tim Januschowski\\
  AWS AI Labs \\
  Krausenstrasse 38, \\
  Berlin, Germany\\
  \texttt{tjnsch@amazon.com} \\
}

\begin{document}

\maketitle

\begin{abstract}
Intermittent demand, where demand occurrences appear sporadically in time, is a common and challenging problem in forecasting. In this paper, we first make the connections between renewal processes, and a collection of current models used for intermittent demand forecasting.
We then develop a set of models that benefit from recurrent neural networks to parameterize conditional interdemand time and size distributions, building on the latest paradigm in ``deep'' temporal point processes.
We present favorable empirical findings on discrete and continuous time intermittent demand data, validating the practical value of our approach.
\end{abstract}
\section{Introduction} \label{sec:introduction}

Large parts of inventory catalogs in e-commerce, manufacturing, and logistics are plagued by ``intermittency,'' where positive demand occurs sporadically in time \cite{bose2017probabilistic,boylan_classification_2008,seeger_bayesian_2016}. 
Concretely, intermittent demand forecasting (IDF) focuses on demand time series, \ie, instantiations of discrete-time nonnegative stochastic processes, which characteristically contain long consecutive runs of zeroes.
Apart from rendering most standard forecasting methods impractical, IDF raises questions on forecast accuracy measures, model selection, model ensembling, etc.
Moreover, intermittent patterns likely appear in slow-moving, high-value items that are critical to production processes, such as spare or specialty parts in aerospace, maritime, and defense \cite{syntetos_intermittent_2011}.
Accurate forecast uncertainty estimates---\ie, having access to forecast distributions---are markedly important in the context of IDF, where inventory control is inherently difficult and consequential.

\begin{figure}[H]
  \makebox[\textwidth][c]{
  \includegraphics[width=1.3\textwidth]{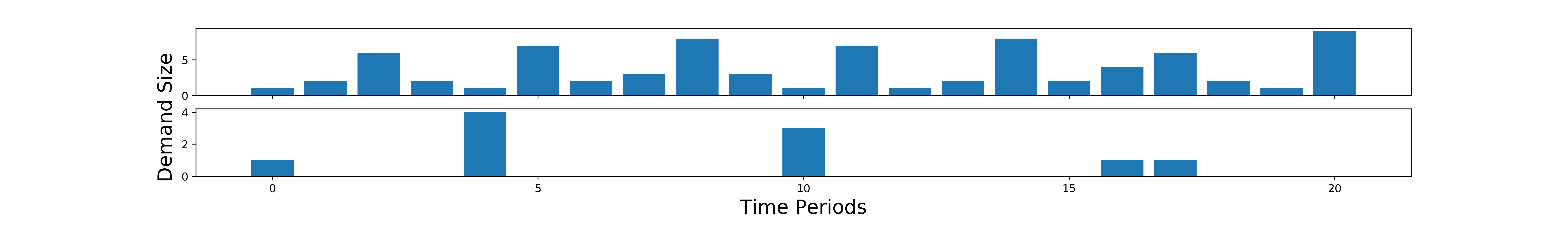}
  }
  \vspace{-1cm}
  \caption{Comparing a non-intermittent count time series (above), and intermittent demand time series (below).}
\end{figure}
\vspace{-.3cm}

The standard approach to IDF, and the first paper to address it, is the method of Croston \cite{croston_forecasting_1972}. 
Croston's method is a rather simple heuristic, forecasts are generated by independently applying exponential smoothing to the amounts of consecutive non-zero demand, and to the times elapsed in between each (\ie, {\em interarrival} or {\em interdemand} times). Towards developing forecast distributions,  models built on a set of constrained statistical assumptions inspired by Croston's method were attempted in \cite{hyndman_forecasting_2008,johnston_forecasting_1996,shenstone_stochastic_2005,snyder_forecasting_2002}. Another important aspect is that, the traidional approach (Croston and others) is \emph{local}, meaning that each time series has its own model and thus no information is shared across time series. Our approach is \emph{global}, learning from all to reach the statistical tail from the sparse signal that each intermittent demand time series supplies. For more discussions on the paradigm shift from the traditional to modern approaches for time series forecast, we encourage interested readers to~\cite{januschowski2019criteria,faloutsos2018forecasting,faloutsos2019classical}

In this work, we first focus on the link between renewal processes and existing approaches to IDF.
Interestingly, although these models have appeared in the backdrop of manufacturing and spare parts (renewals), connections have seldom been drawn. We then point out that many existing IDF models can be seen as instantiations of marked {\em self-modulating} renewal processes in discrete-time. By replacing the conditional distribution function with recurrent neural networks, we can specify a family of flexible models for IDF, akin to some continuous-time processes that have appeared recently in the machine learning literature \cite{du_recurrent_2016,mei_neural_2017}.

Our contributions are three-fold. First, we show the connections between intermittent demand problems and renewal processes, two areas in applied statistics that have dealt with temporal sparsity, and have appeared in the context of planning spare parts inventories in manufacturing. Secondly, we cast Croston-like models as instances of discrete-time conditional (Markov) renewal processes. Through this observation, we introduce recurrent neural networks to recover more flexible ways in which renewal processes are modulated. Finally, we extend our construction in the context where the individual demand events are observed in continuous time. By making connections to continuous time renewal processes and deep temporal point processes, we study how these models apply in the IDF context. 


\section{Preliminaries} \label{sec:preliminaries}

We consider nonnegative integer-valued processes in discrete-time denoted $\{Y_n\}_{n \in \{1, 2, \cdots\}}, Y_n \in \mathbb{N}_{\ge 0}$, where $n$ indexes {\em demand review periods}.
As many $Y_n$ are zero, we introduce the additional index $i$ for periods where $Y_n > 0$. 
More formally, we have the one to one map $\sigma(i) = \min \{n \mid \sum_{m=1}^n [Y_n > 0] \ge i\}$.
We denote {\em interdemand} times $Q_i$, as the number of periods between consecutive demand points, and $M_i = Y_{\sigma(i)}$ are the corresponding demand sizes. 
That is, $Q_i = \sigma(i) - \sigma(i-1)$ taking $\sigma(0) = 0$.
Random variables $M_i, Q_i$, both defined on positive integers, fully determine $Y_n$ and vice versa. 
We will let $\hat{Y}_{n+1}$ refer to a {\em forecast} of $Y_{n+1}$, \ie, an {\em estimator} of the conditional mean $\ex{Y_{n+1} | Y_{1:n}=y_{1:n}}$.

Croston \cite{croston_forecasting_1972} heuristically proposed to produce forecasts by separately performing SES on interdemand times and demand sizes.  
Concretely, 
\begin{subequations}\label{eq:croston_fcast}
\begin{align}
\hat{M}_{i+1} &= \alpha M_{i+1} + (1 - \alpha) \hat{M}_{i}, \label{eq:croston_fcast_1}   \\
\hat{Q}_{i+1} &= \alpha Q_{i+1} + (1 - \alpha) \hat{Q}_{i}. 
\end{align}
\end{subequations}
That is, exponentially weighted moving average (EWMA) are employed separately on the two time series. 

Several authors have taken the same approach---specifying the random process through inter-demand times and sizes---as the basis of underlying statistical models \cite{johnston_forecasting_1996,shenstone_stochastic_2005,snyder_forecasting_2002,snyder_forecasting_2012}.
Croston's original proposed model was taking $M_i, Q_i$ as i.i.d., an assumption clearly at odds with the forecast equations of (\ref{eq:croston_fcast}).
Others proposed to use the EWMA recursion directly to parameterize the conditional distributions of $Q_i, M_i$. 
For example, taking,
\begin{align} \label{eq:mod_croston_model}
M_{i+1} \sim \mathcal{PO}(\hat{M}_i) \qquad Q_{i+1} \sim \mathcal{G}(\hat{Q}_i),
\end{align}
where $\hat{M}_i, \hat{Q}_i$ denote EWMA as defined in (\ref{eq:croston_fcast}), and $\mathcal{PO}, \mathcal{G}$ denote Poisson and geometric distributions respectively.\footnote{All distributions in this work are parameterized by their {\em means}, which we take as the first parameter.}

Here, we observe that taking $Q_i$ i.i.d. equivalently specifies a renewal process in {\em discrete} time.\footnote{This is a rare setting for renewal processes. Yet, some of the earlier works on renewal theory are based on the discrete-time formalism. See, \eg, \cite{feller_introduction_1957}.} 
In the same light, the model specified in (\ref{eq:mod_croston_model}) can be cast as a specific Markov renewal process in discrete time \cite{cinlar_markov_1975}, or a {\em self-modulating} renewal process where the time to the next event depends on a function of previous occurrences. 

\section{Proposed Models} \label{sec:models}
In this section, we propose a novel framework in both discrete and continuous time scenarios.
\subsection{Discrete-Time Deep Renewal Processes} 
We start by extending the generic IDF model, given in (\ref{eq:mod_croston_model}) along two directions.

{\bf Recurrent Neural Networks} The EWMA, or {\em exponential smoothing}, is a forecast {\em function} that often appears in the forecasting literature. 
Apart from favorable properties such as ensuring temporal continuity, and arising as the optimal forecasts to an integrated moving average process; it is just a functional form that has the important property that it admits recursive computation.
Indeed, the same is true for recurrent neural networks which could equivalently be used to replace the EWMA, and serve as an approximator for a more general modulating function for the next event's time. 
Furthermore, the RNN could be conditioned on previous demand {\em sizes} along with interdemand times, at a negligible additional computational cost. 

{\bf Interarrival Time Distributions} To the best of our knowledge, previous studies in IDF have relied on the geometric distribution for modeling interdemand times. 
Analogously to the case of exponential interarrival times in Poisson processes, this imposes the assumption that demand arrivals are completely random.
Nevertheless, intermittent demand series often exhibit autocovariance \cite{willemain_forecasting_1994}.
Here, we use RNNs to parameterize a negative binomial distribution of interdemand times, a choice that affords flexibility in capturing effects such as aging, negative aging (clustering), and quasi-periodicity---all of which can reasonably be expected to manifest in demand observations.

Concretely, we propose the following {\em Discrete-Time Deep Renewal Process} setting
\begin{align*}
&Q_i \sim \mathcal{NN} \left(g_q(\mathbf{h}_i)\right)
&M_i \sim \mathcal{NN} \left(g_m(\mathbf{h}_i) \right)
&&\mathbf{h}_i = LSTM_\Theta (\mathbf{h}_{i-1}, Q_i, M_i), 
\end{align*}
where $\mathbf{h}_i$ denotes the state given by a long short-term memory RNN \cite{hochreiter1997long}, and $\mathcal{NN}$ specifies a generic probability distribution defined on positive integers (possibly with other parameters), and $g_.$ denote suitable projections from the space of vectors $\mathbf{h}_i$ to the space of parameters (means) of the respective distributions.

\subsection{Continuous-Time Renewal Processes} 
Demand data is most often available along a discrete, uniformly sampled grid in time, or {\em demand review periods}. 
For non-intermittent demand, this is a useful abstraction as it provides the required level of accuracy for most inventory control decisions downstream.
However, intermittent demand data often arises as an artifact of discretizing demand point occurrences, localized in continuous time, along a dense temporal grid.
Here, intuitively, discretization error plays a larger role.
Moreover, in more demand forecasting scenarios, exact timestamps are available for individual purchase events---such as in e-commerce.

Let $j$ index individual demand events, such as purchase orders or e-commerce transactions, $\Tilde{Q}_j \in \IR_+$ the interarrival times between such events and $\Tilde{M}_j \in \mathbb{N}$ the demand sizes. 
Analogously to their discrete-time counterparts, continuous time demand occurrences can be modeled simply by assuming a nonnegative continuous distribution for interarrival times, such as the exponential or the gamma distribution, setting 
\begin{align*}
&\Tilde{Q}_j \sim \mathcal{NR} \left(g_q(\mathbf{h}_i)\right)
&\Tilde{M}_j \sim \mathcal{NN} \left(g_m(\mathbf{h}_i) \right)
&&\mathbf{h}_i = LSTM_\Theta (\mathbf{h}_{i-1}, \Tilde{Q}_j, \Tilde{M}_j),
\end{align*}
where $\mathcal{NR}$ is a distribution defined on nonnegative reals.
This construction recovers recurrent marked temporal point processes (RMTPP) \cite{du_recurrent_2016}.\footnote{We should note two differences. RMTPP \cite{du_recurrent_2016} parameterizes the {\em conditional intensity} function of a self-modulating renewal process, rather than the conditional demand time distribution. Also, their paper deals with the case of categorical marks, while we cast demand sizes as marks.}

Our model allows easy implementation, flexibility, and fast computation as self-modulating renewal processes admit likelihood-based inference in time linear in the number of events (as opposed to more general conditional intensity point processes). 
We give a summary of the models tested in this paper in Table~\ref{tab:ct_models}.

\begin{table}[H]
  \centering
  \noindent\makebox[\textwidth]{
    \begin{tabular}{p{5cm} p{3cm} p{3cm} p{3cm}}
      \toprule
      {\bf Model}    & $Q_i$ ($\Tilde{Q}_j$) & $M_i$ ($\Tilde{Q}_j$) & {\bf Recurrence} \\ \midrule
      Static Croston Model \cite{croston_forecasting_1972,snyder_forecasting_2012}   & Geometric & Poisson & N/A (i.i.d.)\\[10pt]
      Modified Croston Model \cite{hyndman_forecasting_2008,shenstone_stochastic_2005}  & Geometric  & Poisson  & EWMA\\[10pt]
      Discrete Time DRP   & Negative Binomial  & Negative Binomial  & LSTM\\[10pt]
      Continuous Time DRP   & Gamma  & Negative Binomial  & LSTM\\[10pt]
      \bottomrule
  \end{tabular}}
  \label{tab:ct_models}
  \caption{Models tested, specified in terms of interarrival time, demand size distributions and the recurrence relation used to parameterize these (conditional) distributions}
\end{table}

\section{Experiments}
We evaluate our model on two real-world datasets. The \texttt{Parts} data set is the standard {\em common task} in IDF \cite{syntetos_intermittent_2011}, and consists of 1046 aligned time series of 50 time steps each, representing monthly sales for different items of a US automobile company. We also report results on the \texttt{UCI} retail data set \cite{chen2012data}, in which timestamps of individual demand events (continuous-time data) are available. 

We implement and train continuous and discrete-time deep renewal process (DRP) models on Apache MXNet \cite{chen2015mxnet}, using MXNet Gluon Adam optimizer for training. 
We use a single hidden layer of 10 hidden units, use the softplus function for mapping the LSTM embedding to distribution parameters.
Most importantly, we learn a {\em global} RNN, \ie, LSTM parameters are shared across different time series.

For evaluation, we use the \emph{quantile loss} to evaluate probabilistic forecasts. For a given quantile $\rho\in(0,1)$, the $\rho$-quantile loss is defined as
\[
\text{QL}_\rho(\obs_t, \widehat{\obs}_t(\rho)) = 2\left[\rho\cdot(\obs_t - \widehat{\obs}_t(\rho))\mathbb{I}_{\obs_t - \widehat{\obs}_t(\rho) > 0} + (1-\rho)\cdot(\widehat{\obs}_t(\rho) - \obs_t)\mathbb{I}_{\obs_t - \widehat{\obs}_t(\rho) \leqslant 0}\right].
\]
We compare our models to static and modified Croston models, where we take forward samples from selected models to generate probabilistic forecasts.
Table~\ref{tab:results} summarizes our results. 

\begin{table}[H]
  \centering
  \noindent\makebox[\textwidth]{
    \begin{tabular}{p{5cm} cccc}
      \toprule
                   & \multicolumn{2}{c}{\textbf{P50QL}} & \multicolumn{2}{c}{\textbf{P90QL}} \\ \midrule
      {\bf Model}                       & \texttt{Parts}  & \texttt{UCI} & \texttt{Parts}  & \texttt{UCI} \\ \midrule
      Static Croston Model \cite{croston_forecasting_1972,snyder_forecasting_2012}      & 3.643    & 12.278 & 2.495    & 7.979 \\
      Modified Croston Model \cite{hyndman_forecasting_2008,shenstone_stochastic_2005}    & 4.218  & 13.453 & 3.507  & 8.371 \\
      Discrete Time DRP                 & {\bf2.835}  & 8.404  & {\bf1.601}  & {\bf4.295} \\
      Continuous Time DRP                  & N/A    & {\bf8.233}  & N/A     & 4.502 \\
      \bottomrule
  \end{tabular}}
  \caption{Results on benchmark datasets}
  \label{tab:results}
\end{table}

DRPs significantly outperform baseline methods on both losses. 
The benefit of using continuous-time DRPs can be tested only on the \texttt{UCI} data set, for which our results are inconclusive. 

\section{Conclusion and Future Direction}
Our paper addresses intermittent demand forecasting using DRPs, a class of models inspired by a connection drawn between intermittent demand problems, renewal processes, and existing deep TPP literature~\cite{turkmen2019fastpoint}.
We find encouraging empirical results, partially presented here, on real-world intermittent demand data.
Validating our approach on a broader array of IDF data sets remains a next step to our work, as well as comparing to other approaches, such as DeepAR (with Negative Binomial likelihood)~\cite{salinas2017deepar}, and non-parametric time series models (NPTS)~\cite{alexandrov2019gluonts}.

\bibliographystyle{plain}
\bibliography{references}

\appendix

\end{document}